\documentclass[10pt,twocolumn]{article}

\usepackage{cvpr}
\usepackage{times}
\usepackage{epsfig}
\usepackage{graphicx}
\usepackage{amsmath}
\usepackage{amssymb}
\usepackage{listings}

\usepackage[pagebackref=true,breaklinks=true,colorlinks,bookmarks=false]{hyperref}

\cvprfinalcopy 

\def\cvprPaperID{****} 

\ifcvprfinal\fi

\title{Automatic Code Generation using Pre-Trained Language Models}

\author{

  Luis Perez \\
  Department of Computer Science \\
  Stanford University \\
  \texttt{luis0@stanford.edu}
  
  \and
  Lizi Ottens  \\
  Department of Computer Science\\
  Stanford University\\
  \texttt{lottens@stanford.edu}
  
  \and
  
  \\
  Sudharshan Viswanathan \\
  Department of Computer Science \\
  Stanford University \\
  \texttt{viswans@stanford.edu}
}

\begin{document}
\maketitle

\begin{abstract}
 Recent advancements in natural language processing \cite{gpt2} \cite{BERT} have led to near-human performance in multiple natural language tasks. In this paper, we seek to understand whether similar techniques can be applied to a highly structured environment with strict syntax rules. Specifically, we propose an end-to-end machine learning model for code generation in the Python language built on-top of pre-trained language models. We demonstrate that a fine-tuned model can perform  well in code generation tasks, achieving a BLEU score of 0.22, an improvement of 46\% over a reasonable sequence-to-sequence baseline. All results and related code used for training and data processing are available on GitHub. \footnote{See shared repository located at \href{https://github.com/kandluis/code-gen}{https://github.com/kandluis/code-gen}.}
\end{abstract}

\section{Introduction}
\label{sec:introduction}
Automating even small parts of software development is an active research area \cite{survey}, with multiple approaches proposed  methods (See Section \ref{sec:introduction}). Succeeding in the automation of even small tasks can save time for countless software engineers, which translates to saved resources across multiple industries. Furthermore, as software continues to eat the world \footnote{See \href{https://www.wsj.com/articles/SB10001424053111903480904576512250915629460}{article} by Marc Andreessen.} and demand for experienced software developers continues to outpace supply, automatic code generation will become increasingly important.

In this paper, we propose a machine learning model to automate the task of writing code by assisting developers in writing individual units of functionality (or ``functions''). Automating code generation can take on many forms, from auto-completing lines of source code to generating lines of source code from comments, generating source code from UI images, or generating unit tests from source code. In this project, we aim to take the initial lines of code (a function signature) along with a doc-string (function documentation) and generate the corresponding function body. In order to do this, we use a pre-trained language model and fine-tune it on a canonical corpus of Python code scraped from GitHub \cite{codesearchnet}.

\section{Background}
\label{sec:background}
A primary challenge in code generation is that it is still an active area of research, with many possible solutions and ongoing investigation \cite{structural_code_modeling}. State of the art solutions have not yet come close to automating basic tasks software engineers perform on a daily basis.

\subsection{Traditional Code Completion}
The most traditional and well-known approach used by multiple IDEs across a range of languages simply consists of token completion based on structured information obtained from static analysis of code. For example, when a developer types a sequence of characters, the system will attempt to find near-matching strings corresponding to function definitions and propose completing these function calls. Similarly, for object methods, on the typing of the accessor token (such as ``->'' or ``.''), the IDE will propose auto-completing different methods belonging to the object.

The biggest drawback of these approaches is that they lack true understanding of the programmers intent, and also lack context relating to the surrounding code other than that from heuristics by the tool's developers.

\subsection{Using Machine Learning for Code Search}
Another approach taken in multiple papers in the literature \cite{codesearchnet} involves framing the problem as a code search problem. Rather than trying to generate code or complete the code that the developer is making, we can re-frame the problem as one of searching for relevant pre-existing snippets. This is the primary approach we take in three of our baseline models.

\subsection{Using Machine Learning for Code Generation}
Other more novel approaches from literature \cite{structural_code_modeling} are typically applied to restricted language domains, and have massive complexity in evaluation results, etc. Specifically, while pre-trained models are trained on free-form language data, programming languages often utilize non-natural variable names, function names, and syntax with more structure \cite{structural_code_modeling}. Work in this area has focused on creating more structured models that take advantage of specific architectures \cite{treegen}. In \cite{treegan}, the authors work to first decompose the input sequence of text tokens for the context into a tree-like structure. Other approaches involve restricting the output of the model to a context-free grammar (CFG) or domain-specific language (DSL) \cite{grammar_based}. A code generation model’s output must adhere to a very specific form in order to be syntactically correct.

In this paper, we instead focus on taking a different approach. As has been demonstrated by ever-increasing sizes of language models, we focus on improving the performance on the code prediction task by making use of pre-trained language models that are then fine-tuned on code. 

\subsection{Dataset and Feature}
\begin{figure}[!ht]
    \centering
    \includegraphics[scale=0.1]{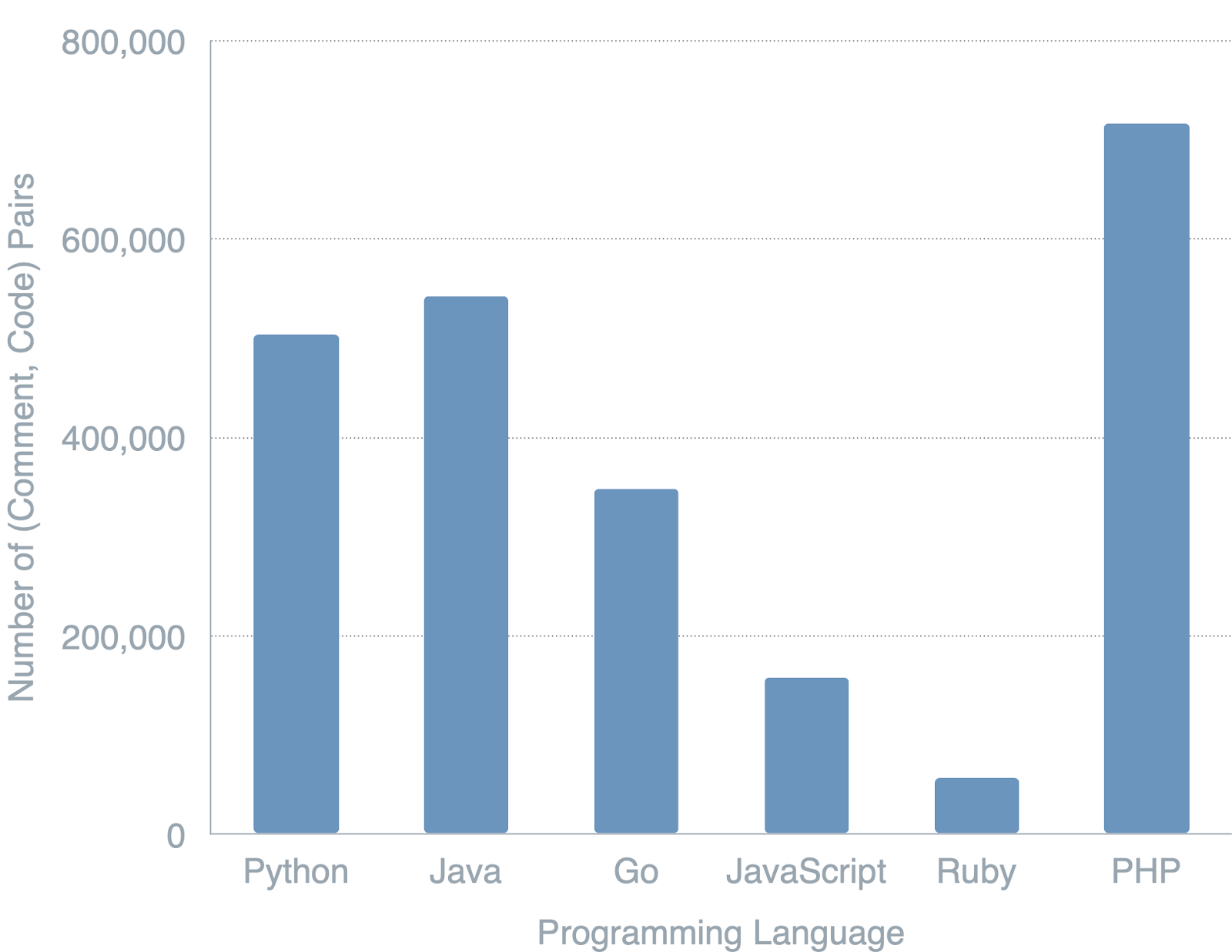}
    \includegraphics[scale=0.1]{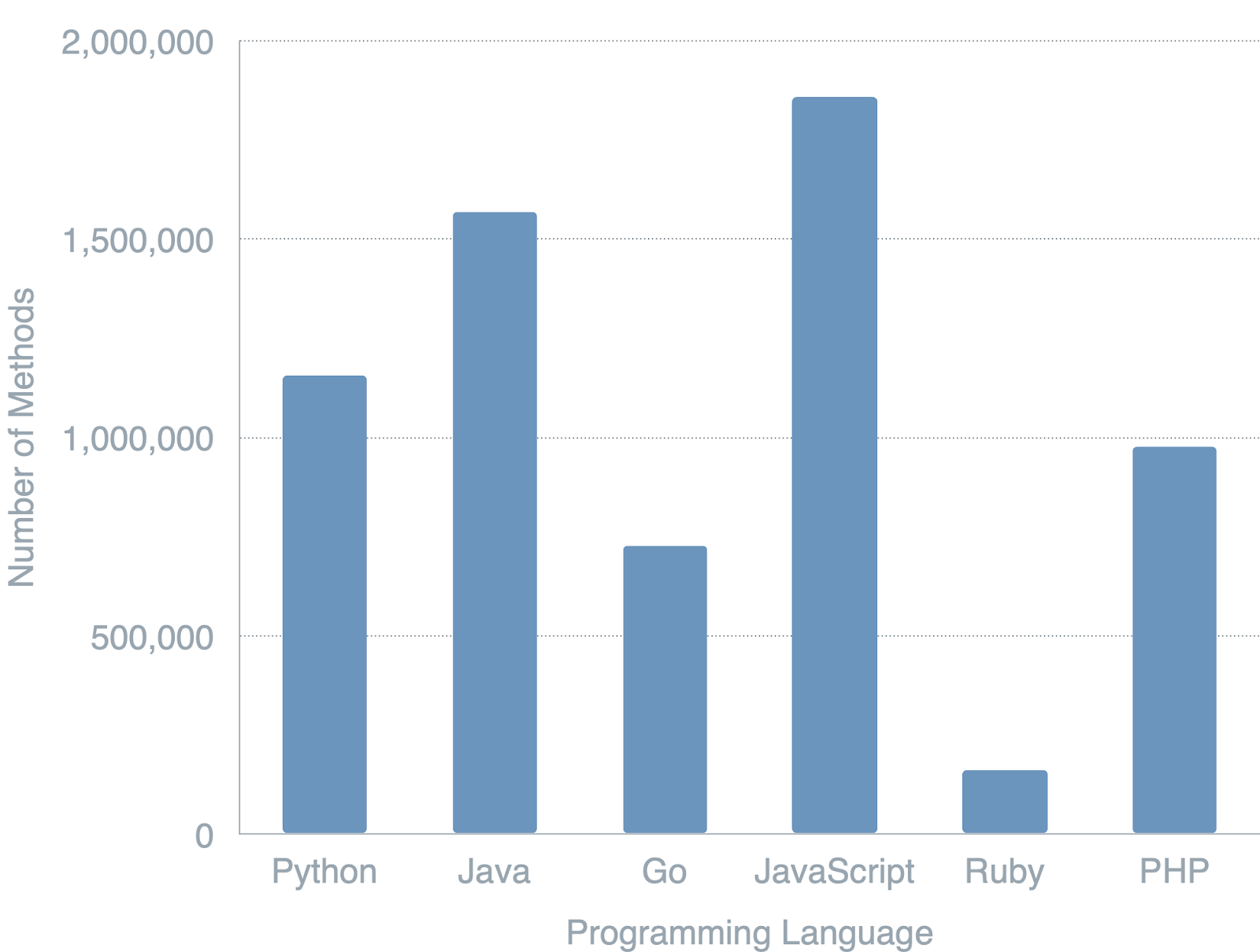}
    \caption{Histogram of the the number of (comment, code) pairs available in our dataset, as well as the number of unique function methods for each language.}
    \label{fig:dist_methods}
\end{figure}

In this project, we are leveraging the \href{https://github.com/github/CodeSearchNet}{CodeSearchNet dataset} \cite{codesearchnet}. The dataset consists of 2 million (comment, code) pairs from open source libraries, ranging in languages from Python to Javascript, PHP, Java, Go and Ruby. Median code-length consists of 60-100 text tokens, with 95\% code-length of up to 350 tokens. Median documentation length consists of 10 text tokens. The distributions of methods and (comment, code) pairs across programming language are visualized in Figure \ref{fig:dist_methods}.

We restrict our dataset to samples in the Python programming language rather than the others available. Focusing on Python, there are over 1M methods and approximately 500k (comment, code) pairs that make up our dataset. We make this decision both for practical and modeling reasons. From a practical perspective, restricting to a reasonably-sized dataset focused on a single-language domains permits for more thorough ablation studies. From a modeling perspective, we belief that transfer learning from natural language to a programming language such as Python is an easier task to accomplish.

\section{Methodology}
In this section, we explain our methodology for multiple experiments and baselines proposed as well as details on the training data and distribution.

\subsection{CodeSearchNet Models}
\begin{figure}[!ht]
    \centering
    \includegraphics[scale=0.15]{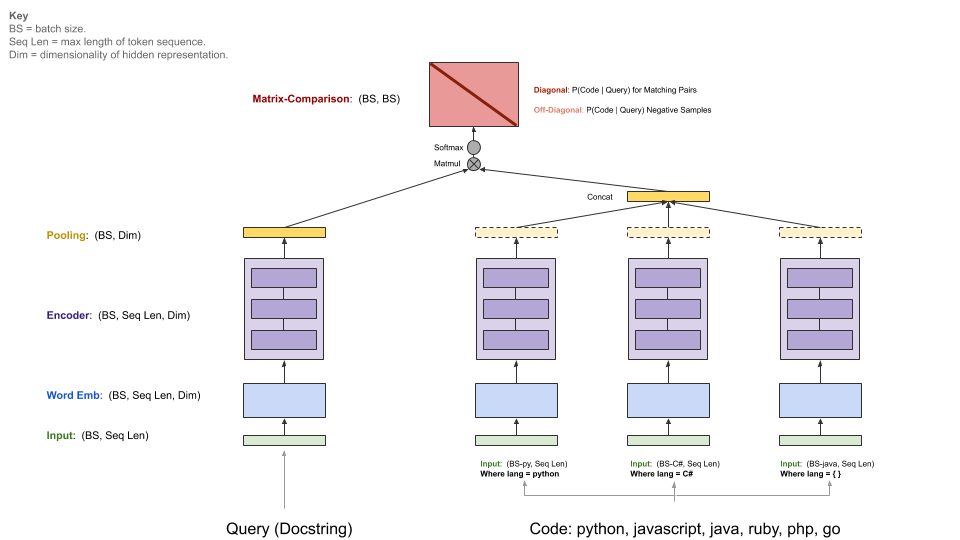}
    \includegraphics[scale=0.15]{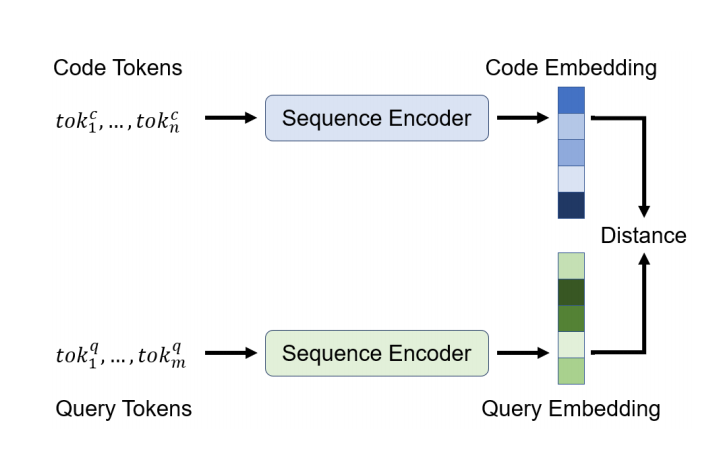}
    \caption{General CodeSearchNet architecture for all of our baselines. Each language is processed through different encoder mechanisms. The query encoder is shared (an NLP encoder), and the purpose of the CodeSearchNet tasks is to retrieve the most relevant code snippets subject to the natural language query.}
    \label{fig:codesearchnet_arch}
\end{figure}

Figure \ref{fig:codesearchnet_arch} explains the general architecture of the baseline models from the CodeSearchNet task. We successfully trained and evaluated two baselines: Neural-Bag-Of-Words and an RNN-based baseline. See Section \ref{sec:results}.

Generally speaking, the baselines models take as input examples of (comments, code) pairs and learn to retrieve a specific code snippet. Each programming language has its own encoder network (see three columns to the right in Figure \ref{fig:codesearchnet_arch}), which are tasked with encoding a set of candidate code snippets. They are then combined through a dot product operation with the embedding generated by the query (docstring) encoder to produce a matrix comparison. 

The matrix diagonal serves as the scores of each query doc string/code snippet. Through this methodology, these baseline models are able to extract meaningful information and learn a joint distribution over the query and comment pairs. We train these models as a baseline since we believe they will be useful in the downstream task of code generation. The models are trained on the following loss function:

\begin{align}
    -\frac{1}{N}\sum_i \log \left( \frac{\exp(E_c(\textbf{c}_i^T)E_q(\textbf{d}_i))}{\sum_{j} \exp(E_c(\textbf{c}_j^T)E_q(\textbf{d}_j))}  \right)
    \label{eq:code_search_net_loss}
\end{align}

\subsection{From Scratch RNN Models}
The above baseline is useful only in the sense that it would allow our system to find pre-existing code snippets which might be relevant to the developer. Since our goal is rather to make novel code, we propose a different baseline based on a more traditional sequence-to-sequence model.

In this case, we use a traditional RNN architecture which takes as input individual characters. The reason we take this approach is to circumvent the need to learn word-level embeddings. Furthermore, we hypothesize that making use of entire words, from NLP models, will actually harm the performance of the model for code generation. The primary reason for this being that most of the syntax involved in writing code does not generally map directly to the English language. 
Concretely, we encode each character present in the training data as a 1-of-k encoding (one-hot encoding) and feed them into an RNN one at a time. Our output will be a k-dimensional output vector corresponding to a probability distribution over the entire set of characters.

\begin{figure}[!ht]
    \centering
    \includegraphics[scale=0.25]{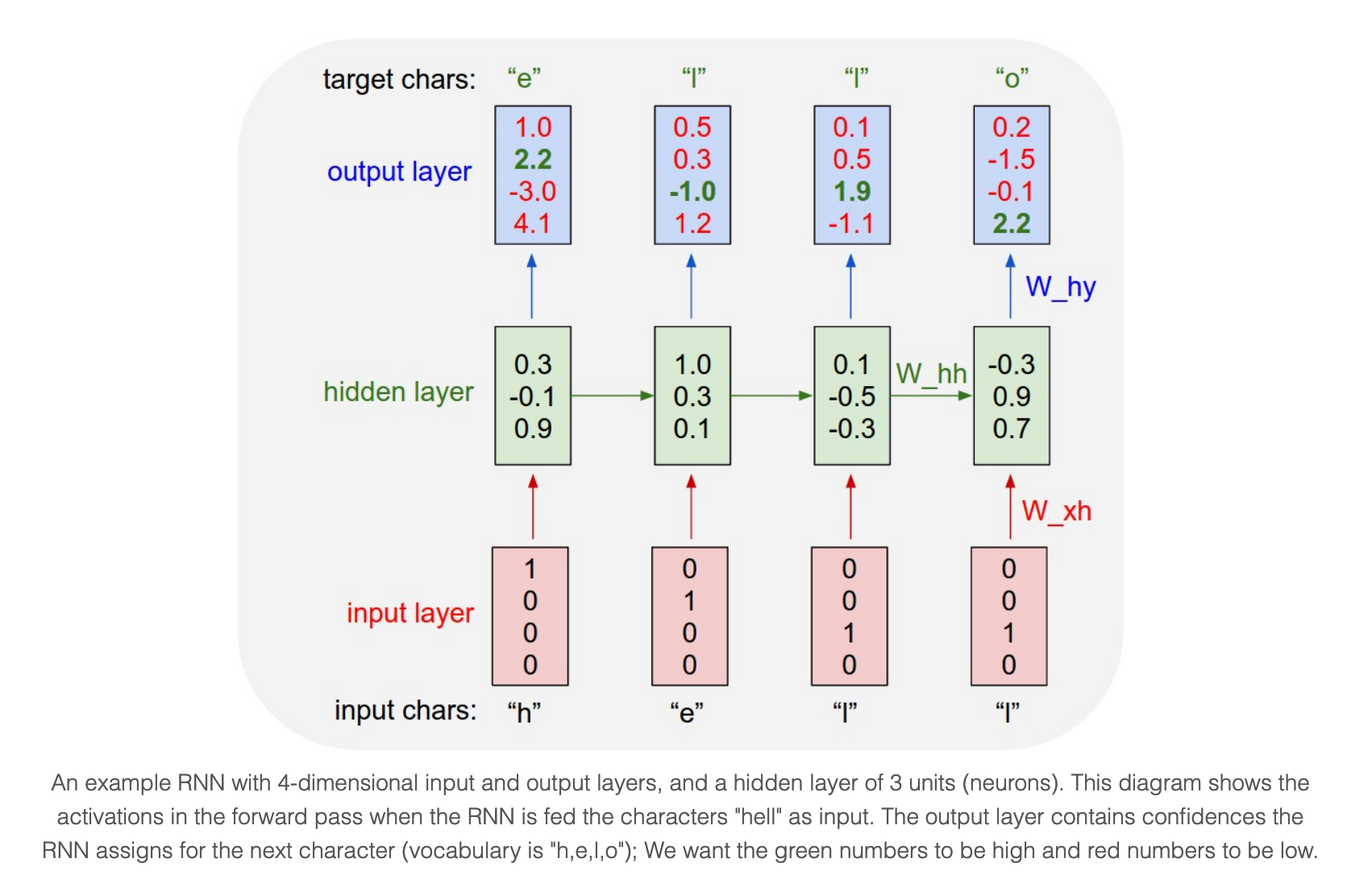}
    \label{fig:char-rnn-arch}
\end{figure}

For the model architecture, we sweep over multiple types of RNN cells, including LSTM, RNN, and GRU. We find the best performing model to consists of an LSTM-based model using a  hidden state size of $128$ with two hidden layers in the internal RNN cell. Our training takes place using sequences of $50$ characters, sampled at random from our input code. Given a sequence from $i$ to $i + 50$, the model is trained to predict the sequence from $i + 1$ to $i + 51$. This means we have a many-to-many sequence model (See Figure \ref{fig:char-rnn-arch}). We use batch size of $50$ and train for a total of $50$ epochs. 

To avoid issues with gradient explosion and stabilize training, we make liberal use of gradient-clipping. In particular, we clip all gradients to an absolute size of $5$. 

We sweep over learning rates and find that a started learning rate of $0.002$ with an exponentially decaying schedule appears to perform best as measured by a held-out validation set. We use a decay rate of $0.97$ per epoch. We also experiment with the use of dropout, but find little impact on final performance.

\subsection{Fine-Tuned Pre-Trained Large Language Models}
Our final approach relies on the use of pre-trained language models. We fine tune our code generator using the small GPT-2 model with 117 million parameters. Using such a large backbone and continuing to fine tune allows us to generate synthetic code samples with even higher quality, treating programming languages as another specific domain alongside encyclopedia articles, news or books.

\begin{figure}[!ht]
    \centering
    \includegraphics[scale=0.2]{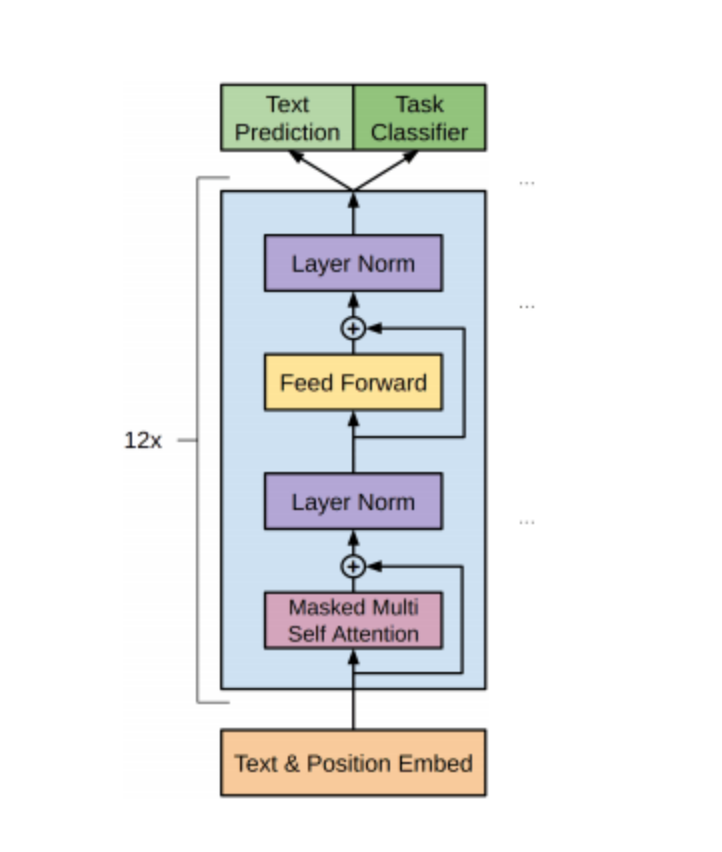}
    \caption{Decoder-Only Architecture used by GPT-2.}
    \label{fig:gpt2_architecture}
\end{figure}

The general architecture of the GPT-2 model consists of a sequence-to-sequence predictive task based on the transformer architecture \cite{attentionisallyouneed} \cite{gpt2}. However, it consists solely of the 12-layer decoder-only, as visualized in Figure \ref{fig:gpt2_architecture}. Each layer has 12 independent attention heads, leading to 144 distinct attention patterns. By making use of an attention-based framework, the model is more adept at dealing with long-range dependencies. This is because the attention mechanism allows the model to focus on the encoding of any of the input sequence tokens. 

\section{Results}
\label{sec:results}
CodeSearchNet provides a good starting point as we are able to train different models on the input code streams. We trained a simple LSTM model as well as a neural bag of words model on a combination of all the available (code, documentation) pairs. For details on these simple baselines, please see Appendix Section \ref{sec:codesearchnet_results}.

\subsection{Code Generation with Char-RNN}
As both of the above baselines focus on understanding and extracting useful embeddings for our overall task, our primary baseline consists of a straight-forward sequence-to-sequence model. Given that code typically does not consist of English words and can instead have quite a varied syntax, our baseline model is a model which uses character level embedding, so it is character aware \cite{charrnn}.

\begin{figure}[!ht]
    \centering
    \includegraphics[scale=0.15]{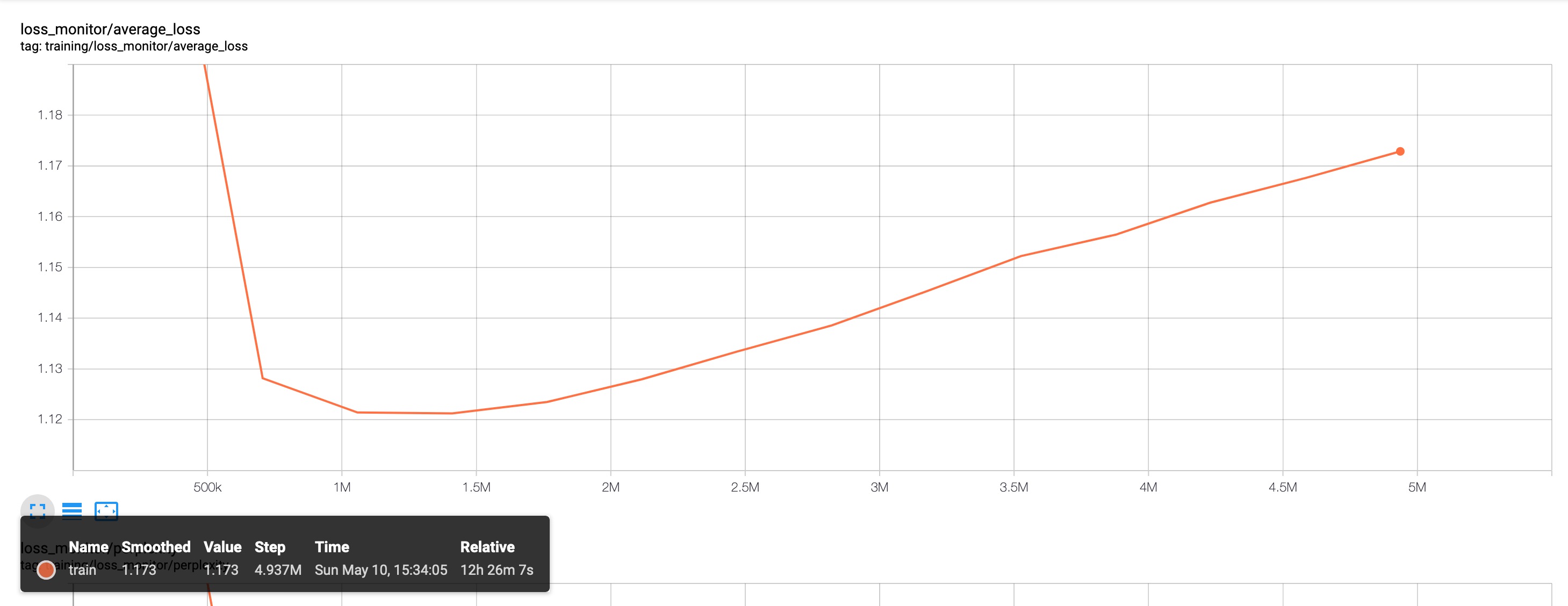}
    \includegraphics[scale=0.15]{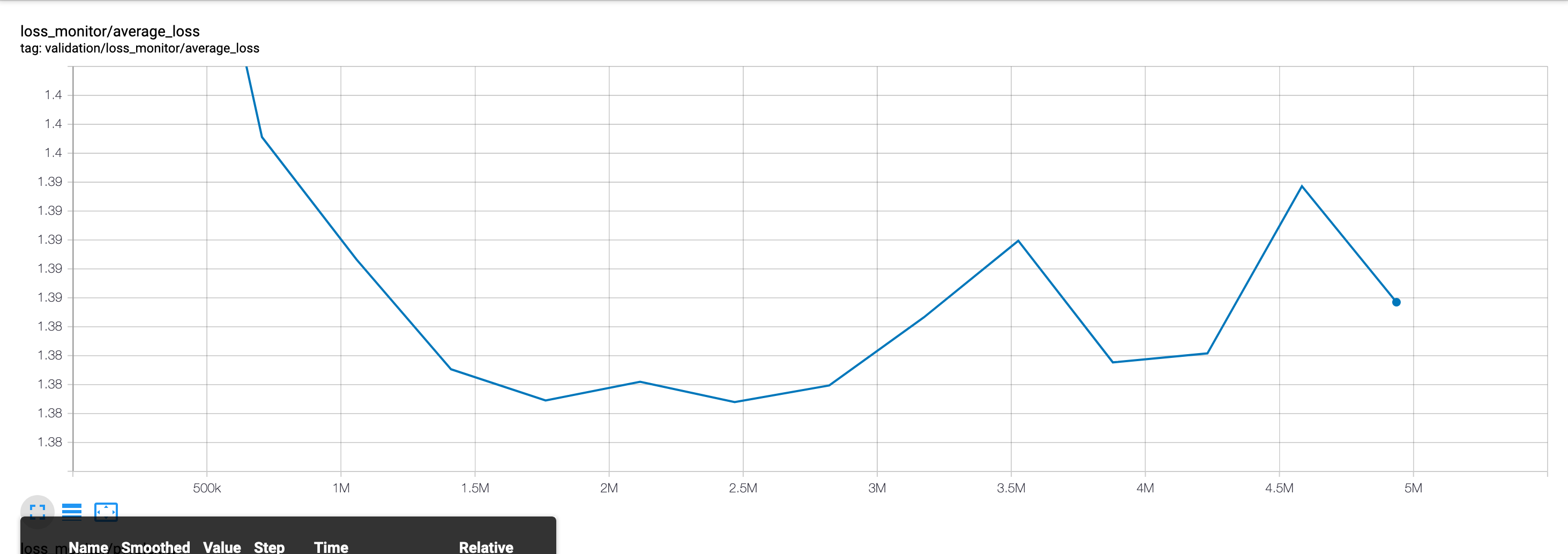}
    \caption{Training and Validation Losses on the Baseline Char-RNN Model. This is the cross-entropy loss over 128 predicted character sequence.}
    \label{fig:char_rnn_loss}
\end{figure}

\begin{figure}[!ht]
    \centering
    \includegraphics[scale=0.15]{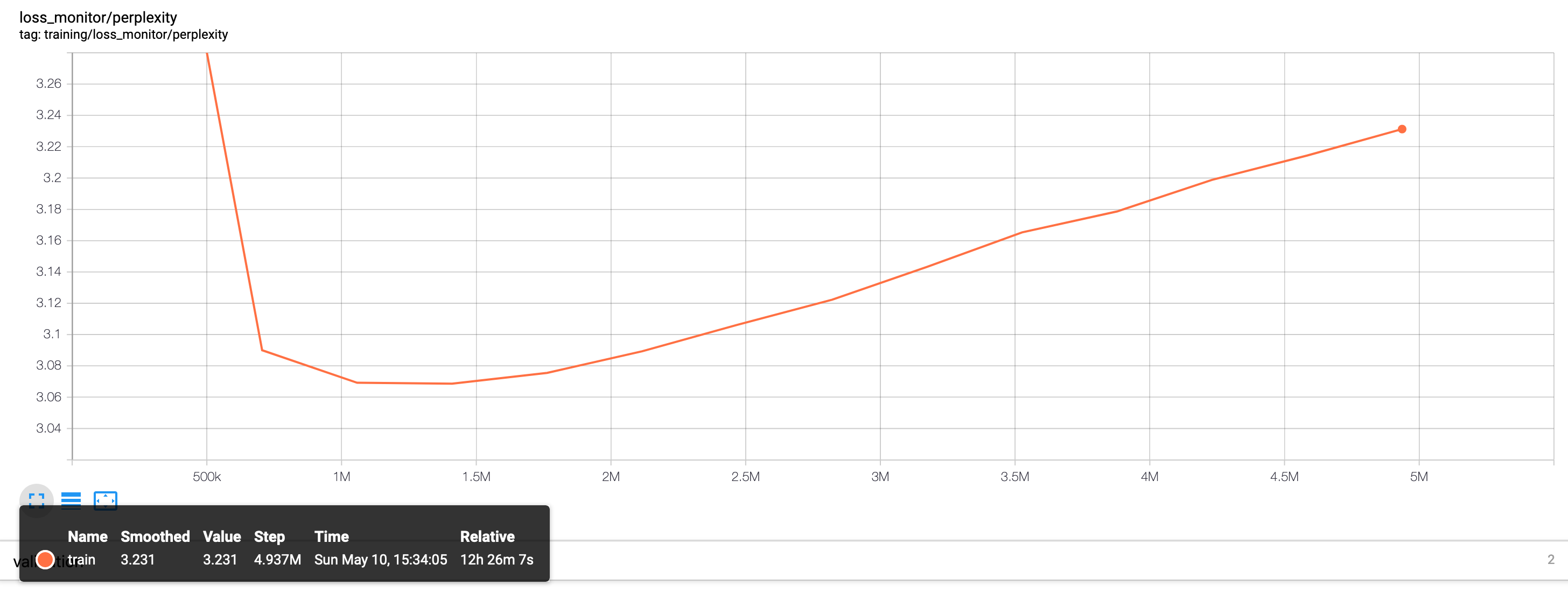}
    \includegraphics[scale=0.15]{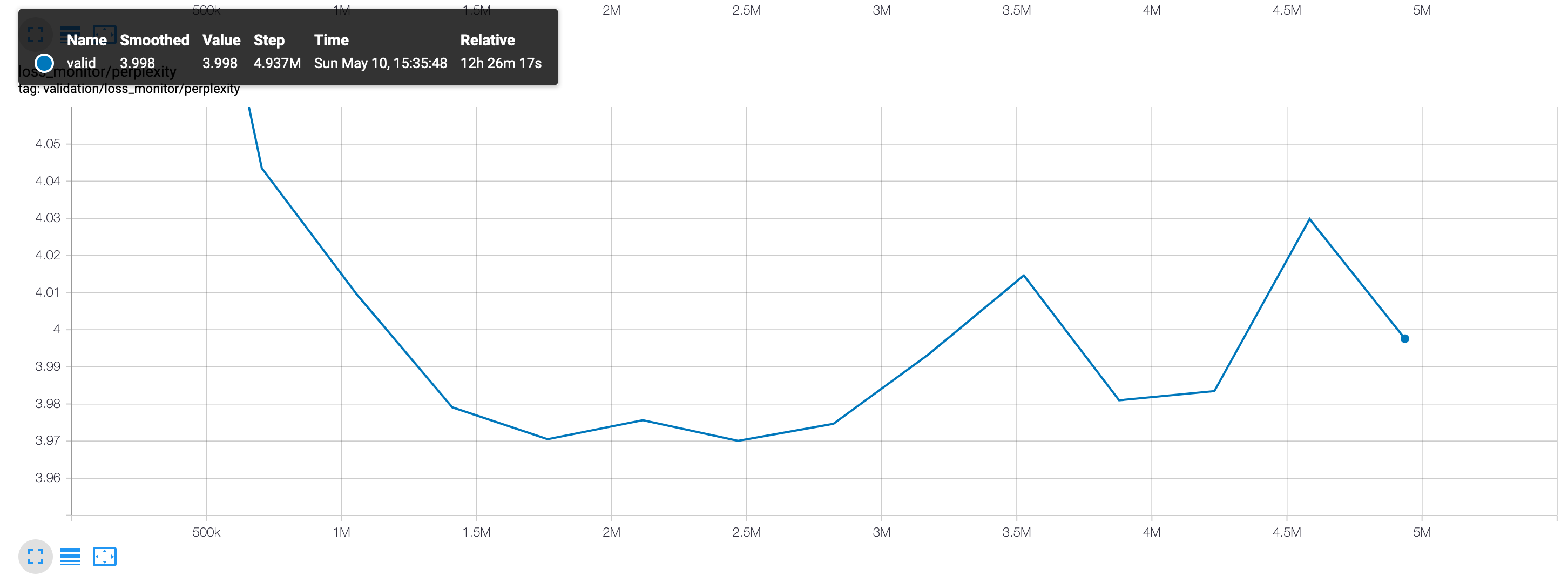}
    \caption{Training and Validation Perplexity on the Baseline Char-RNN Model. This is the cross-entropy loss over 128 predicted character sequence.}
    \label{fig:char_rnn_perplexity}
\end{figure}

Due to computational constraints, we train only on the Python subset of the data and only on 10\% of the total data available. For the char-rnn model \cite{charrnn}, this corresponds to around 50MB of raw text, or 78,357,395 characters with 1,618 distinct symbols. Figure \ref{fig:char_rnn_loss} shows the training and validation losses on the model. The loss is simply a softmax loss on the 1,618 characters for a sequence of length 128 (the model is trained on sequences of length 128 by default). Figure \ref{fig:char_rnn_perplexity} shows the perplexity, or the amount of meaningful information encoded. 

We include a sample generated from the best performing model for reference (See Section \ref{sec:background} in Appendix). A hyper-parameter tuning of learning rate and batch side for a total of 20 epochs has final measured performance as shown in Table \ref{table:charr_rnn_sweep_label}.

\begin{table}[!ht]
\centering
\begin{tabular}{lllll}
\hline
\textbf{\begin{tabular}[c]{@{}l@{}}Batch\\ Size\end{tabular}} &
  \textbf{\begin{tabular}[c]{@{}l@{}}Starter\\ Learning\\ Rate\end{tabular}} &
  \textbf{\begin{tabular}[c]{@{}l@{}}Regularization\\ Weight\end{tabular}} &
  \textbf{\begin{tabular}[c]{@{}l@{}}BLEU Score\\ on Train\end{tabular}} &
  \textbf{\begin{tabular}[c]{@{}l@{}}BLEU Score\\ on Eval\end{tabular}} \\ \hline
\multicolumn{1}{|l|}{64}  & \multicolumn{1}{l|}{0.02}   & \multicolumn{1}{l|}{0.1}  & \multicolumn{1}{l|}{0.022}          & \multicolumn{1}{l|}{0.012}          \\ \hline
\multicolumn{1}{|l|}{64}  & \multicolumn{1}{l|}{0.02}   & \multicolumn{1}{l|}{0.01} & \multicolumn{1}{l|}{0.023}          & \multicolumn{1}{l|}{0.019}          \\ \hline
\multicolumn{1}{|l|}{64}  & \multicolumn{1}{l|}{0.002}  & \multicolumn{1}{l|}{0.1}  & \multicolumn{1}{l|}{0.034}          & \multicolumn{1}{l|}{0.028}          \\ \hline
\multicolumn{1}{|l|}{64}  & \multicolumn{1}{l|}{0.002}  & \multicolumn{1}{l|}{0.01} & \multicolumn{1}{l|}{0.037}          & \multicolumn{1}{l|}{0.012}          \\ \hline
\multicolumn{1}{|l|}{64}  & \multicolumn{1}{l|}{0.0002} & \multicolumn{1}{l|}{0.1}  & \multicolumn{1}{l|}{0.09}           & \multicolumn{1}{l|}{0.073}          \\ \hline
\multicolumn{1}{|l|}{64}  & \multicolumn{1}{l|}{0.0002} & \multicolumn{1}{l|}{0.01} & \multicolumn{1}{l|}{0.094}          & \multicolumn{1}{l|}{0.014}          \\ \hline
\multicolumn{1}{|l|}{128} & \multicolumn{1}{l|}{0.02}   & \multicolumn{1}{l|}{0.1}  & \multicolumn{1}{l|}{0.024}          & \multicolumn{1}{l|}{0.021}          \\ \hline
\multicolumn{1}{|l|}{128} & \multicolumn{1}{l|}{0.02}   & \multicolumn{1}{l|}{0.01} & \multicolumn{1}{l|}{0.021}          & \multicolumn{1}{l|}{0.013}          \\ \hline
\multicolumn{1}{|l|}{128} & \multicolumn{1}{l|}{0.002}  & \multicolumn{1}{l|}{0.1}  & \multicolumn{1}{l|}{0.033}          & \multicolumn{1}{l|}{0.029}          \\ \hline
\multicolumn{1}{|l|}{128} & \multicolumn{1}{l|}{0.002}  & \multicolumn{1}{l|}{0.01} & \multicolumn{1}{l|}{0.038}          & \multicolumn{1}{l|}{0.011}          \\ \hline
\multicolumn{1}{|l|}{128} & \multicolumn{1}{l|}{0.0002} & \multicolumn{1}{l|}{0.1}  & \multicolumn{1}{l|}{\textbf{0.117}} & \multicolumn{1}{l|}{\textbf{0.093}} \\ \hline
\multicolumn{1}{|l|}{128} & \multicolumn{1}{l|}{0.0002} & \multicolumn{1}{l|}{0.01} & \multicolumn{1}{l|}{0.113}          & \multicolumn{1}{l|}{0.034}          \\ \hline
\end{tabular}
\label{table:charr_rnn_sweep_label}
\end{table}

\subsection{Code Generation with GPT-2}
We have been working with the publicly available small GPT-2
model with 117 million parameters. We trained using the small GPT-2 model for 100,000 mini-batch iterations with a batch size of 2. We have included some sample code that our model generated directly in the report. Qualitatively, our model generates code which is far more reasonable than our baseline. The generated code is novel, as verified by doing n-gram overlap analysis between the generated code and the training dataset. We also note that the model learns appropriate understanding of Python syntax, with uses of if-statements, function and method calls, as well as regularly commented code. For full output, see Appendix Section \ref{sec:sample_code}.

We observed that the idea of using Byte Pair encoding as used in GPT-2 is a much better strategy to generate code than just using characters, while of course the size of the models itself has a very observable effect in generating Python-like code.

\begin{figure}[!ht]
    \centering
    \includegraphics[scale=0.4]{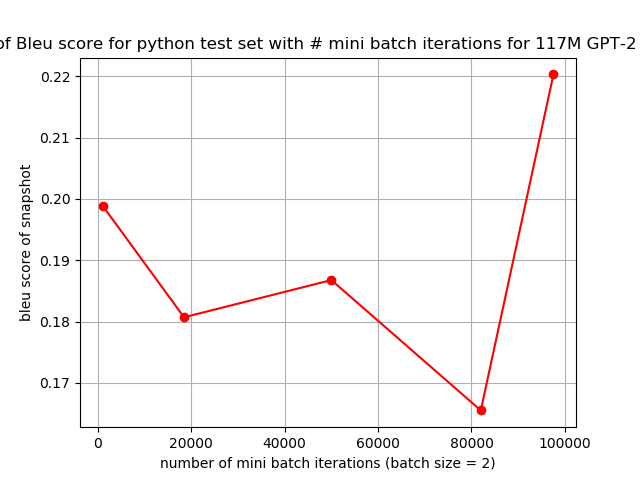}
    \caption{BLEU Score During Training of GPT-2 Based Model for Python Code Generation}
    \label{fig:gpt_bleu_score}
\end{figure}

Overall, the GPT-2 model quickly achieves performance that's much better than the baseline. Continued training of the model shows that our BLEU score performance will continue to increase, as seen in Figure \ref{fig:gpt_bleu_score}

\section{Conclusions}
In this paper, we explore the problem of automatically completing a function from the given function signature and human-readable documentation. We find the best performing model to be a fine-tuned version GPT-2, a transformer-based NLP model which is trained to generate natural text on an extremely large dataset. Despite the fact that our model focuses specifically on code rather than natural language, we hypothesize that it is able to treat programming language as another specific domain alongside the encyclopedia articles, news or books that its backbone has been trained on. We are able to achieve a BLEU score of 0.22, improving our baseline by >40\%.

\section{Contributions}
All team member contributed equally to this project. Baselines from the CodeSearchNet models for code search were trained and tuned by Luis Perez and Sudharshan Viswanathan. Data analysis and understanding of the features (including histograms, distribution of tokens, and other data insights) was primarily performed by Lizi Ottens.

Training of the baseline char-rnn model, as well as analysis of results and discussion was contributed primarily by Luis Perez. Fine-tuning and training with the small and medium GPT-2 models was primarily explored and analyzed by Lizi Ottens and Sudharshan Viswanathan.

All written submissions were co-written by all three authors. 

\bibliographystyle{unsrt}
\bibliography{references}

\newpage
\onecolumn
\section*{Appendix and Figures}
\subsection{CodeSearchNet Results}
\label{sec:codesearchnet_results}

The Neural Bag of Words and LSTM CodeSearchNet baselines both report metrics in the same fashion. Below, we show the training curves, which correspond to the loss in Equation \eqref{eq:code_search_net_loss}.

Additionally, given that the baselines models for CodeSearchNet focus on code snippet retrieval, we also report the achieved mean reciprocal rank. The MRR is a statistic measure for evaluating any process that produces a list of possible responses to a sample of queries, ordered by probability of correctness. The reciprocal rank of a query response is the multiplicative inverse of the rank of the first correct answer: 1 for first place, $\frac{1}{2}$ for second place, $\frac{1}{3}$ for third place and so on. The mean reciprocal rank is the average of the reciprocal ranks of results for a sample of queries, as in Equation \eqref{eq:mrr}.

\begin{align}
    \text{MRR} = \frac{1}{|Q|}\sum_{i=1}^{|Q|} \frac{1}{\text{rank}_i}
    \label{eq:mrr}
\end{align}

\subsubsection{Neural Bag of Words Baselines}
\begin{figure}[!ht]
    \centering
    \includegraphics[scale=0.1]{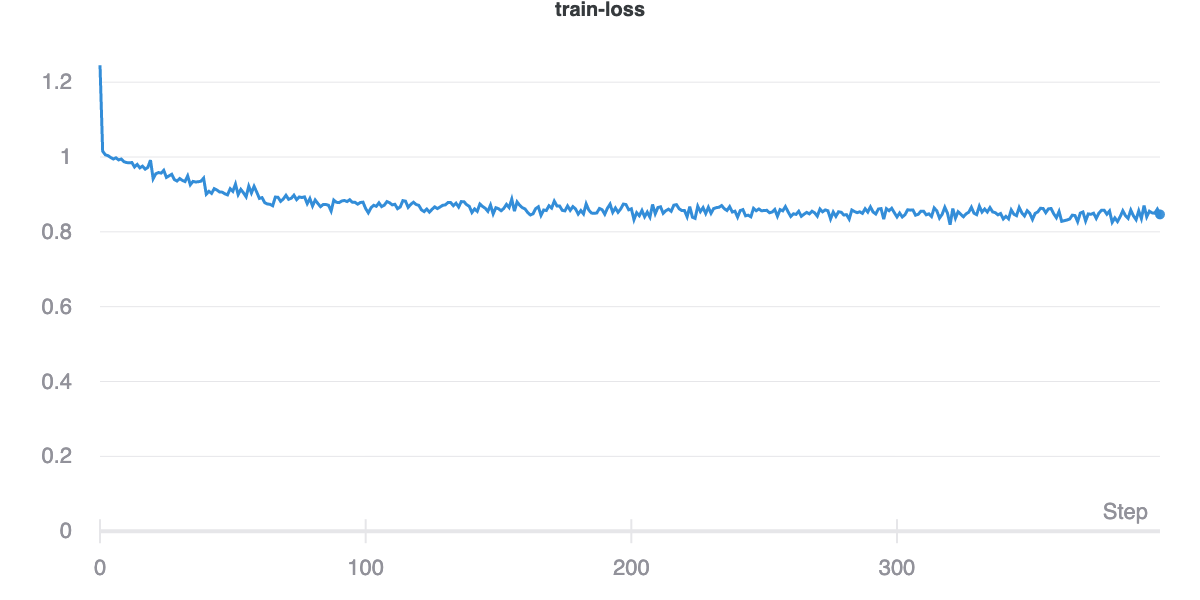}
    \includegraphics[scale=0.1]{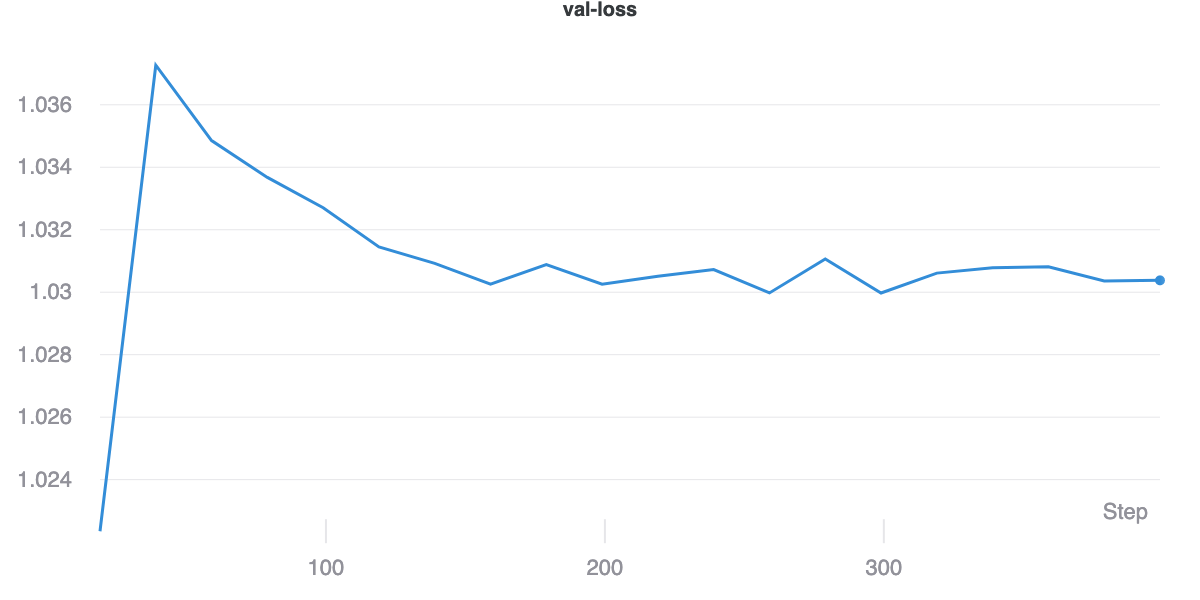}
    \caption{Training and Validation losses for the Neural Bag of Words model in CodeSearchNet.}
    \label{fig:train_valid_neural_bow_model}
\end{figure}

\begin{figure}[!ht]
    \centering
    \includegraphics[scale=0.15]{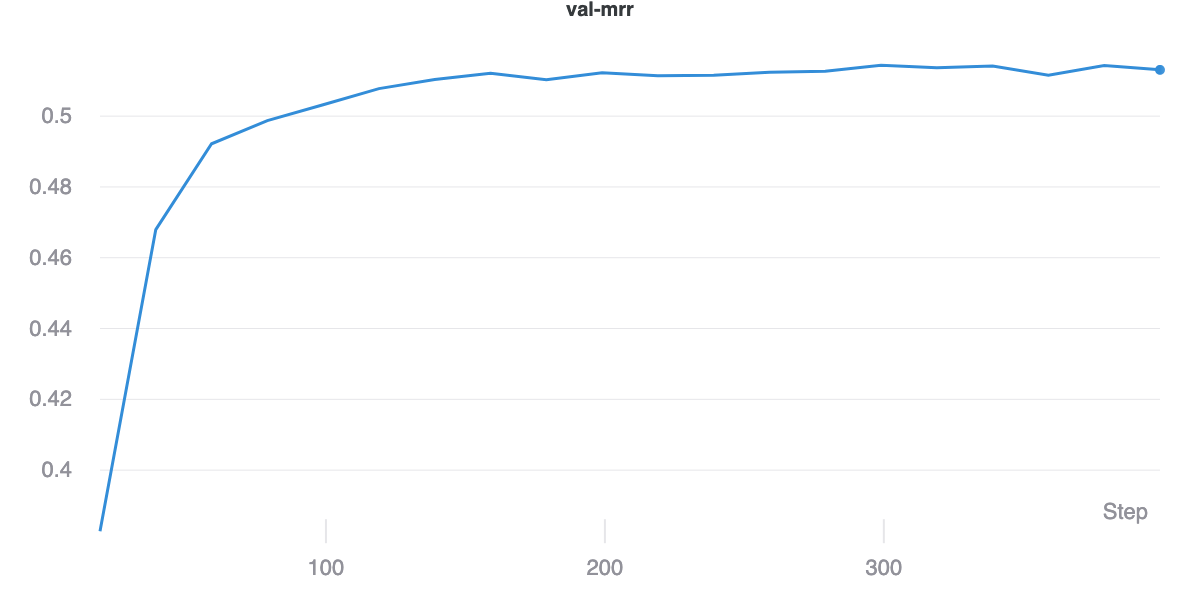}
    \caption{MRR on validation set for the baseline neural bag of words model in the CodeSearchNet Challenge.}
    \label{fig:val_mrr_neuralbow}
\end{figure}

This baseline consists of a simple encoder architecture which takes as input bag-of-words representation of the code and using a single neural network encodes these token representation into an embedding \cite{codesearchnet}. This baseline actually performs the best, achieving the lowest overall training and validation losses (see Figure \ref{fig:train_valid_neural_bow_model}) as well as the highest MRR on the validation set (See Figure \ref{fig:val_mrr_neuralbow}).

\subsubsection{Bi-directional RNN Model}
\begin{figure}[!ht]
    \centering
    \includegraphics[scale=0.1]{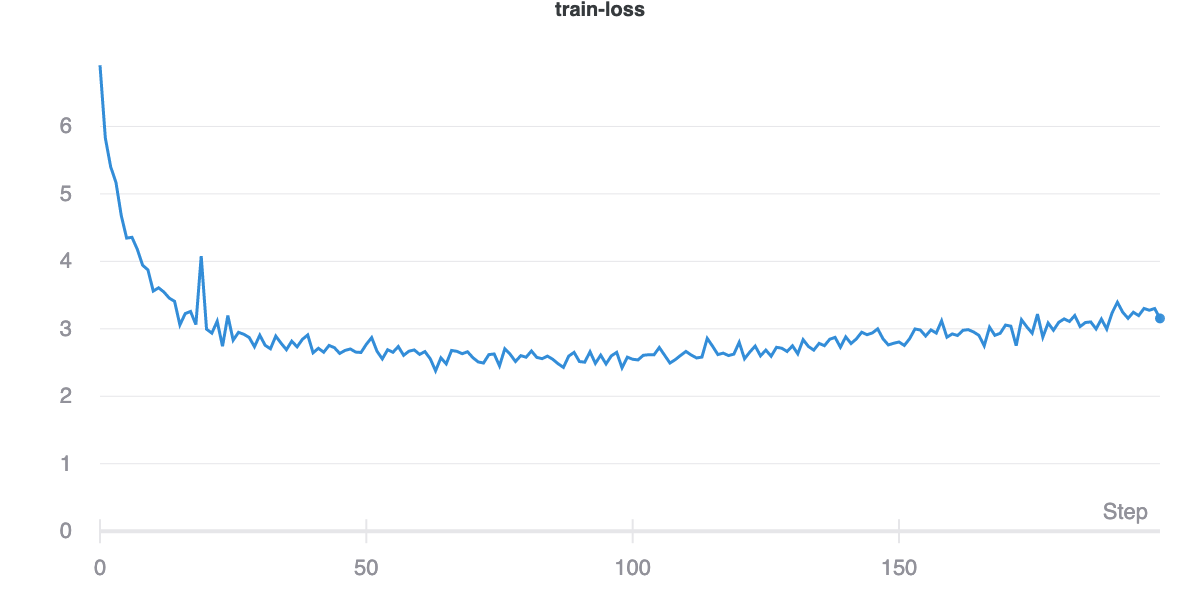}
    \includegraphics[scale=0.1]{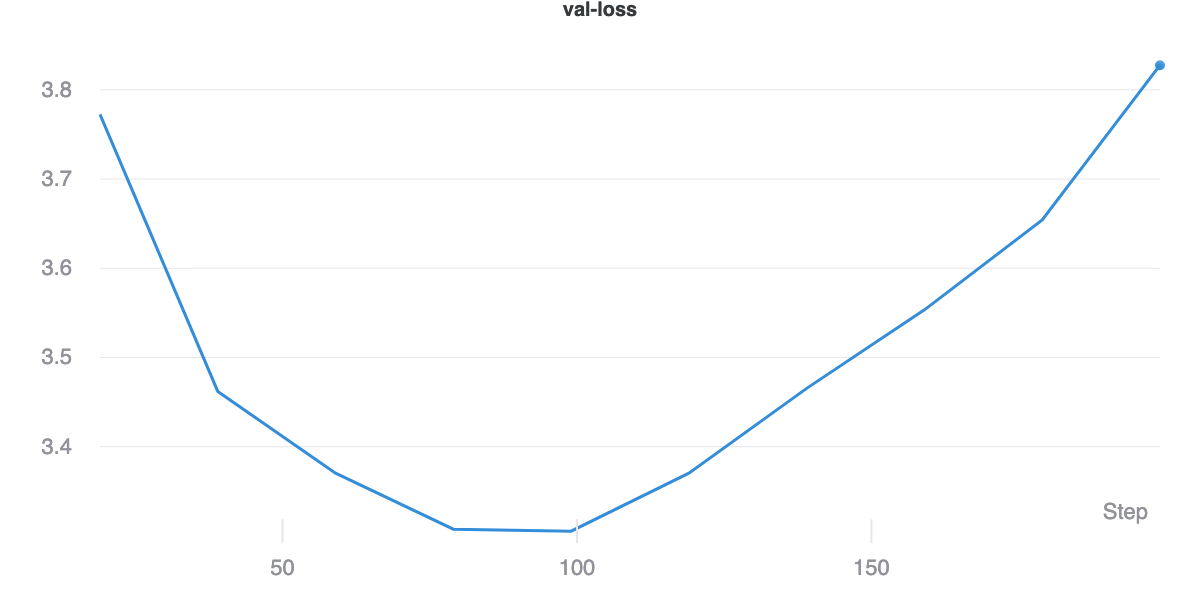}
    \caption{Training and Validation losses for the RNN model in CodeSearchNet.}
    \label{fig:train_valid_neural_rnn_model}
\end{figure}

\begin{figure}[!ht]
    \centering
    \includegraphics[scale=0.15]{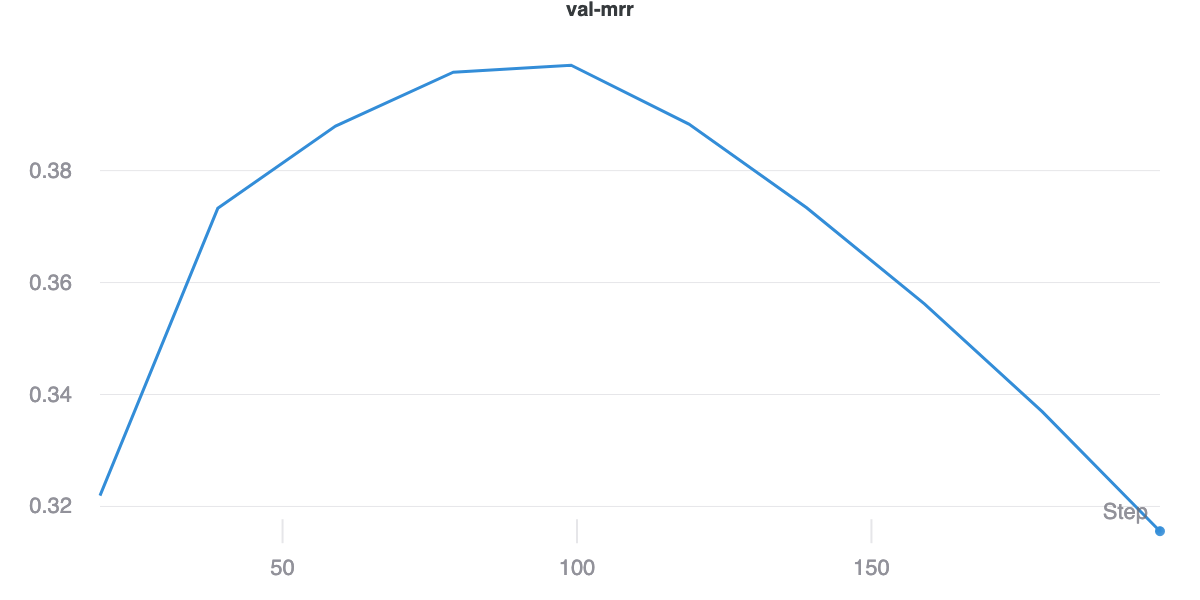}
    \caption{MRR on validation set for the baseline RNN in the CodeSearchNet Challenge.}
    \label{fig:val_mrr_rnn}
\end{figure}

In this model, we employ the GRU cell \cite{gru_cell} to summarize the input sequence. This baseline performs significantly worse, suffering from what appears to be obvious over-fitting. In Figure \ref{fig:train_valid_neural_rnn_model}, we can see that while the training loss appears to plateau, the validation loss begins quickly climbing. While this behavior does not appear to affect the overall MRR achieved on the validation set, it is still clear that the model performs worse than the bag of words baseline as per Figure \ref{fig:val_mrr_rnn}.

\subsection{Example Code}
\label{sec:sample_code}
\begin{lstlisting}[language=Python,caption=Sample generated using Char-RNN model.]
Downloads Dailymotion videos by URL. (x or pillares (if macks), style as a bool to you extner method instruxError,
       to oio instance of that Seir to be two categorical String to mandation :attr:`ColumnserverName(
                zr, j)
        dimensed_source
        also = ``axis``. Element. This repr_dures a substimcle of

        """
        code = item
    return self.filename
            restroxig, self.get_channels():
    """Get the function as firmap_list {1} and a :attracted_coordint: value of Time
    end_aes:
            target_ext = int(cmd_dict):
            # In data applicate horinad(string): the channel for decoded collfile
            Runnee network number of element. What you's associates py ch of List does request. Channels represented should be a TargetExcept the more notes

        ``index``.
def vert_event_notify_channel(self, infna=1)
                    elif trying.format(
        comps + ',' % (random) + 1,
            H.data:
            if list + lookbing.get('get_dict.get']}
\end{lstlisting}

\begin{lstlisting}[language=Python,caption=Sample generated using small (117M) GPT-2 model after 40000 iterations.]
input_snippet = def semilocal_linear_trend_transition_matrix(autoregressive_coef):
  """Build the transition matrix for a semi-local linear trend model.""",
output_code = 
  # Create the matrix
  # (autoregressive_linear_trend_matrix) where :param semilocal_linear_trend_matrix:
  #                 (n_samples) is length of the semi-sparse linear regression matrix;
  #                 i, j = semilocal_linear_trend

    # Create the columns
    columns = np.asarray(columns)
    if auto_correlated_trend_matrix:
        column = np.asarray(columns)

    # Create the indices for the columns
    i = column.min()
    j = column.max()
    # Ensure the columns have uniformity and the axis position
    assert hasattr(
        column, "sigma")
    assert hasattr(axes, "axis")

    # Create the columns
    i = columns.copy()
    j = columns.copy()

    # Ensure the indices for the indices for the columns
    ax = ColumnSetAxes(i, j)
    axes.set_axes(ax)

    # Create the columns
    ax.set_xlabel(self.default_label)
    ax.set_ylabel(self.default_ylabel)
    axes.set_title(self._title)
    axes.text(u"Subsim"]

    # Create the column index
    self.get_index(axis = columns, cols = cols)

    # Create the columns if not already provided by this function
    if not columns:
        if auto_correlated_trend_matrix:
            ax.columns = columns
        else:
            if i < 0 or j < 0 or i+1 <= i
\end{lstlisting}

\subsubsection{Figures}

\end{document}